\documentclass[10pt, a4paper]{article}

\usepackage[final]{lrec2026} 

\usepackage{amsmath,cite,url}
\usepackage{natbib}
\usepackage{booktabs}
\usepackage{multirow}
\usepackage{tcolorbox}
\tcbuselibrary{skins}
\usepackage{enumitem}
\usepackage{makecell}
\usepackage[dvipsnames]{xcolor}

\title{ArtistMus: A Globally Diverse, Artist-Centric Benchmark for Retrieval-Augmented Music Question Answering}

\name{Daeyong Kwon, Seungheon Doh, Juhan Nam} 

\address{Graduate School of Culture Technology, KAIST \\
         Daejeon, South Korea \\
         \{ejmj63, seungheondoh, juhan.nam\}@kaist.ac.kr\\}

\abstract{
Recent advances in Large Language Models (LLMs) have transformed open-domain question answering, yet their effectiveness in music-related reasoning remains limited due to sparse music knowledge in pretraining data. While music information retrieval and computational musicology have explored structured and multimodal understanding, few resources support factual and contextual music question answering (MQA) grounded in artist metadata or historical context.
We introduce MusWikiDB, a vector database of 3.2M passages from 144K music-related Wikipedia pages, and ArtistMus, a benchmark of 1,000 questions on 500 diverse artists with metadata such as genre, debut year, and topic. These resources enable systematic evaluation of retrieval augmented generation (RAG) for MQA.
Experiments show that RAG markedly improves factual accuracy—open-source models gain up to +56.8 percentage points (pp; Qwen3 8B: 35.0$\rightarrow$91.8), approaching proprietary performance. RAG-style fine-tuning further boosts both factual recall and contextual reasoning, yielding strong improvements on both in-domain and out-of-domain benchmarks. MusWikiDB also yields +6 pp higher accuracy and 40\% faster retrieval than the general Wikipedia corpus.
We release MusWikiDB and ArtistMus to advance research in music information retrieval and domain-specific QA, establishing a foundation for retrieval augmented reasoning in culturally rich domains such as music.
 \\ \newline  }

\begin{document}

\maketitleabstract

\section{Introduction}



\begin{figure*}[!ht]
\begin{center}
\includegraphics[width=\linewidth]{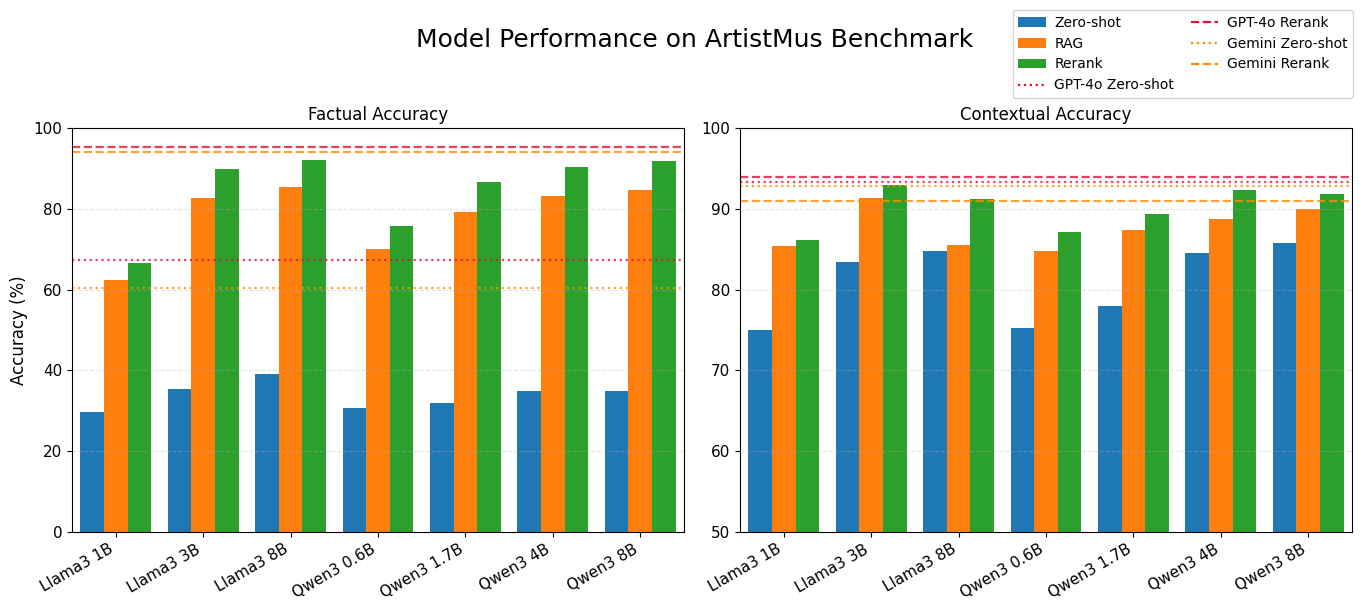}
\caption{Performance comparison of open-source LLMs on the ArtistMus benchmark under three settings (Zero-shot, RAG, and Rerank). The dashed lines denote closed-source models (GPT-4o and Gemini 2.5 Flash) evaluated in the same conditions. The Rerank strategy consistently yields the highest factual and contextual accuracy across all open models, narrowing the gap with proprietary systems.}
\label{fig:main_figure}
\end{center}
\end{figure*}

Large Language Models (LLMs) have demonstrated impressive capabilities across diverse domains, yet their effectiveness in music-related tasks remains limited due to sparse music-specific coverage in pretraining data. This limitation is particularly evident in Music Question Answering (MQA), where users seek detailed information about artists, genres, instruments, and music history—knowledge that general-purpose LLMs often lack or hallucinate \citep{ramoneda2024role}.

Traditional approaches to domain adaptation rely on fine-tuning LLMs with domain-specific data \citep{jeong2024fine,sahoo2024large}. However, this faces significant challenges: securing high-quality training data, escalating computational costs, and the persistent difficulty of continuously updating model knowledge as new music emerges. Retrieval Augmented Generation (RAG) \citep{lewis2021rag} offers a compelling alternative by grounding model outputs in external knowledge bases, enabling dynamic access to specialized information without retraining.

Consider a user querying: \textit{"Which jazz artists influenced Thad Jones's move to Copenhagen?"} General-purpose LLMs often fail at such queries, either hallucinating biographical details or providing generic responses lacking specific career contexts. While fine-tuning could address this, it requires expensive retraining for each knowledge update (e.g., new album releases, career milestones). Moreover, music knowledge is inherently dynamic and culturally situated, making static parametric storage inadequate. This motivates a retrieval-based approach where models can access up-to-date, verifiable information while maintaining the flexibility to reason over retrieved context.

Despite RAG's success in domains such as medicine \citep{miao2024integrating} and law \citep{wiratunga2024cbr}, its application in music remains underexplored due to the absence of music-specific retrieval resources. Existing knowledge bases like Wikipedia lack the density and structure needed for effective MQA. Similarly, current music benchmarks \citep{weck2024muchomusic,yuan2024chatmusician,ramoneda2024role} focus primarily on music theory or multimodal understanding, leaving a critical gap in evaluating artist-centric knowledge—information central to everyday music discovery and consumption \citep{lee2010analysis,doh2024music}.

To address these gaps, we introduce two foundational resources for music-domain question answering. First, MusWikiDB is a specialized knowledge base containing 3.2M passages from 144K music-related Wikipedia pages, covering seven categories: artists, genres, instruments, history, technology, theory, and forms. Unlike general Wikipedia corpora, MusWikiDB is optimized for music retrieval with domain-specific indexing and passage segmentation. Second, ArtistMus is a benchmark of 1,000 questions covering 500 globally diverse artists (spanning 163 countries), with rich metadata including genre, debut year, and topic category.


Using these resources, we conduct systematic experiments showing that retrieval-augmented approaches markedly improve music QA performance. Open-source models equipped with MusWikiDB-based retrieval achieve substantial gains in factual accuracy, performing on par with or exceeding proprietary systems. Moreover, RAG-style training on \textit{(context, question, answer)} triples further enhances factual recall and contextual reasoning. The framework generalizes well to out-of-domain benchmarks and yields more accurate and efficient retrieval than the general Wikipedia corpus.\footnote{Code and data at \url{https://github.com/DaeyongKwon98/ArtistMus}}

Our contributions are as follows:
\begin{itemize}
\item We release \textbf{MusWikiDB}, a comprehensive music-specific vector database containing 3.2M passages from 144K music-related Wikipedia pages, designed for retrieval-augmented music QA.
\item We present \textbf{ArtistMus}, a benchmark of 1,000 validated questions on 500 globally diverse artists, with rich metadata enabling fine-grained analysis.
\item Using MusWikiDB and ArtistMus, we apply RAG to the music domain, achieving substantial improvements in factual accuracy and enabling open-source models to approach proprietary performance.
\end{itemize}

\section{Related Works}

\subsection{Music Question Answering Benchmarks}

Question Answering (QA) is a core task in Information Retrieval, involving the provision of accurate answers to given questions by retrieving relevant information from document collections \citep{allam2012question}. While open-domain QA systems operate over broad knowledge bases \citep{rajpurkar2016squad,karpukhin2020dense}, domain-specific QA targets specialized fields such as medicine \citep{pal2022medmcqa}, law \citep{chalkidis2021lexglue}, or music, where both documents and queries are confined to domain-specific knowledge.

Several recent benchmarks have emerged to evaluate LLM performance on music-related tasks. MuChoMusic \citep{weck2024muchomusic} presents 1,187 audio-based multiple-choice questions assessing musical knowledge and reasoning. MusicTheoryBench \citep{yuan2024chatmusician} contains 372 expert-validated questions evaluating advanced music theory knowledge. TrustMus \citep{ramoneda2024role} comprises 400 questions across four domains—People, Instruments \& Technology, Genres/Forms/Theory, and Culture \& History—derived from The Grove Dictionary Online \citep{sadie2001new}. ZIQI-Eval \citep{li2024music} offers 14,000 comprehension tasks spanning 10 major topics and 56 subtopics.

However, a critical gap exists in existing benchmarks: the inadequate representation of artist-centric metadata—information crucial for everyday music listening contexts \citep{lee2010analysis,doh2024music}. Current text-only QA benchmarks lack comprehensive coverage of details that listeners frequently seek, such as complete discographies, collaboration networks, creative evolution across albums, and career achievements. This disparity between current MQA capabilities and practical information demands motivates the development of artist-focused evaluation resources.

\subsection{Retrieval Augmented Generation for Domain Adaptation}

Domain adaptation of LLMs traditionally relies on fine-tuning with domain-specific data \citep{jeong2024fine,sahoo2024large,satterfield2024fine}. However, this approach faces challenges: securing high-quality training data, escalating computational costs with model scale, and difficulties in continuously updating model knowledge.

Retrieval Augmented Generation (RAG) \citep{lewis2021rag} offers an alternative by augmenting LLMs with external knowledge retrieval. Rather than relying solely on parametric memory, RAG retrieves relevant passages from external databases during inference to ground responses in factual context. Prior work demonstrates that RAG improves factual accuracy, reduces hallucination, enables transparent source verification, and provides greater scalability and cost-efficiency compared to fine-tuning alone \citep{borgeaud2022improving,yasunaga2022deep,wang2023self}.

While RAG has been successfully applied in general-purpose \citep{ram2023context} and domain-specific settings such as law \citep{wiratunga2024cbr} and medicine \citep{miao2024integrating}, its application in the music domain remains underexplored. Music QA requires both factual recall (e.g., artist names, compositions, release dates) and contextual reasoning (e.g., cultural influences, historical relations), making it an ideal testbed for evaluating RAG's effectiveness in culturally rich, knowledge-intensive domains.

Recent work has also explored RAG-style fine-tuning, where models are trained on (context, question, answer) triples rather than traditional (question, answer) pairs \citep{shuster2021retrieval}. This approach explicitly teaches models to integrate supporting passages, potentially improving their ability to leverage external evidence. However, systematic evaluation of this strategy within specialized cultural domains like music has been limited.

Our work addresses these gaps by introducing music-specific resources for RAG-based MQA and providing empirical evaluation of both retrieval-based inference and RAG-style fine-tuning in this domain.



\section{Dataset}

Our resource design follows three key principles to ensure long-term utility for the research community. First, reproducibility: we document all collection and processing steps to enable replication and extension to other music knowledge sources. Second, diversity: both MusWikiDB and ArtistMus prioritize global representation, moving beyond Western-centric coverage typical of existing resources. Third, rich annotation: metadata such as genre, debut year, and topic enable fine-grained analysis.

\begin{table*}[t]
\centering
\resizebox{0.9\textwidth}{!}{%
\begin{tabular}{@{}llll@{}}
\toprule
\textbf{Category} & \textbf{Page} & \textbf{Section} & \textbf{Text} \\ \midrule
Instruments & Wind instrument & Families & Modern brass instruments generally come in ... \\ \midrule
Genres & Music of Tanzania & Singeli & Singeli is a music genre that originated in ... \\
\bottomrule
\end{tabular}%
}
\caption{Examples from MusWikiDB\_raw. It includes the category, page title, section name, and corresponding text.}
\label{tab:muswikidb_raw}
\end{table*}

\begin{table}[t]
\centering
\resizebox{\columnwidth}{!}{%
\begin{tabular}{@{}lrr@{}}
\toprule
\textbf{Category} & \textbf{Page Count} & \textbf{Avg. Tokens} \\ \midrule
Music History    & 40{,}692 & 312 \\
Artist           & 31{,}847 & 528 \\
Genres           & 28{,}553 & 324 \\
Instruments      & 27{,}490 & 316 \\
Musical Forms    & 7{,}681  & 331 \\
Music Theory     & 6{,}055  & 356 \\
Music Technology & 2{,}071  & 301 \\ \midrule
\textbf{Total}   & \textbf{144{,}389} &  \textbf{331} \\ \bottomrule
\end{tabular}%
}
\caption{Number of unique Wikipedia pages and average token count per category in MusWikiDB\_raw.}
\label{tab:unique_pages_per_category}
\end{table}

\begin{table}[!t]
\centering
\resizebox{\linewidth}{!}{
\begin{tabular}{@{}lcc@{}}
\toprule
\multicolumn{1}{l}{} & \textbf{MusWikiDB} & \textbf{Wikipedia Corpus} \\ \midrule
\# Pages             & 0.14M     & 3.2M \\
\# Passages          & 3.2M      & 21M \\
Total tokens         & 0.36B     & 2.1B \\
Vocab Size           & 7.5M      & 21.5M \\ \bottomrule
\end{tabular}
}
\caption{MusWikiDB and Wikipedia Corpus~\citep{karpukhin2020dense} statistics.}
\label{tab:db_statistics}
\end{table}

\subsection{MusWikiDB}

To address the lack of a music-specific vector database for RAG in MQA, we developed MusWikiDB. We began by collecting music-related content from Wikipedia across seven categories: \textit{artists, genres, instruments, history, technology, theory}, and \textit{forms}. These categories were selected to cover a broad spectrum of music knowledge, providing a well-rounded foundation for answering music-related questions. The data collection process began with seven representative Wikipedia pages as entry points. From each entry page, we followed the hyperlinks embedded within the page to gather the first layer of linked pages, denoted as depth 1. Next, for every page at depth 1, we further collected the pages linked within them, which we refer to as depth 2. Finally, the same procedure was applied once more to the depth 2 pages, retrieving their linked pages as depth 3. In this way, the crawling process expanded from the initial seven root pages outward, capturing increasingly detailed subtopics and related information up to a depth of 3. We split the content into wikipedia sections such as \textit{history, background}, and \textit{early life}. We then removed sections shorter than 60 tokens to ensure the remaining text had enough context for meaningful retrieval. Table~\ref{tab:muswikidb_raw} shows an example of the source data of MusWikiDB (MusWikiDB\_raw). Table~\ref{tab:unique_pages_per_category} summarizes the number of Wikipedia pages and the average token count across the seven categories.

To construct the RAG database from MusWikiDB\_raw, the text was segmented into passages of up to 256 tokens with a 10\% overlap between adjacent segments, based on the ablation study in Section~\ref{sec:retriever_configuration}, in order to preserve contextual continuity. For first-stage retrieval, we employed BM25~\citep{robertson1994some}, a classical yet highly effective algorithm for text relevance ranking, to build an efficient index for MusWikiDB. This enabled fast retrieval of relevant information during RAG-based inference, thereby improving both the accuracy and contextual relevance of the generated answers. The resulting MusWikiDB serves as a scalable and up-to-date knowledge base, enhancing RAG performance in MQA tasks and enabling the system to address complex, domain-specific music-related questions with greater precision and context.

Table~\ref{tab:db_statistics} compares our proposed MusWikiDB with the Wikipedia corpus~\citep{karpukhin2020dense}. While MusWikiDB contains fewer pages (0.14M vs 3.2M) and has a smaller vocabulary size (7.5M vs 21.5M), it consists exclusively of music-specialized text information.







\begin{tcolorbox}[float, floatplacement=t, colback=gray!5, colframe=black!50,
                  title=\textbf{Example Questions of ArtistMus}, left=5pt]

\textbf{Metadata:} \\
\textit{Artist}: Robyn \\
\textit{Country}: Sweden \\
\textit{Debut Year}: 1989 \\
\textit{Genre}: Pop \\
\textit{Topic}: Artistry

\tcblower

\textbf{Factual Question:} \\
"What vocal range does Robyn have?" \\
A. Alto \\
B. Soprano \\
C. Tenor \\
D. Baritone \\
\textbf{Answer:} B

\tcbline

\textbf{Contextual Question:} \\
"How did Robyn's approach to music change while writing her eighth studio album, Honey, according to the text?" \\
A. Robyn explored music that was hypnotic and lacked a traditional structure. \\
B. Robyn aimed to replicate the success of her previous hits. \\
C. Robyn focused on creating more mainstream pop hits. \\
D. Robyn returned to her R\&B roots with a focus on tidy pop songs. \\
\textbf{Answer:} A
\end{tcolorbox}

\subsection{ArtistMus}

While previous benchmarks have mainly addressed general music understanding, none has focused on music metadata, particularly the artist, which is crucial in music listening contexts~\citep{lee2010analysis, doh2024music}. Therefore, we introduce ArtistMus, a multiple-choice benchmark dataset on music artists, to evaluate the performance of LLMs in artist-related QA using artist-specific data from MusWikiDB.

\subsubsection{Data Collection}
We selected music artists from MusWikiDB\_raw whose infoboxes contained both \textit{genre} and \textit{year} information. The sections from their Wikipedia pages were then ranked by frequency, and the top five topics—\textit{biography}, \textit{career}, \textit{discography}, \textit{artistry}, and \textit{collaborations}—were chosen to capture the broadest range of artist-related content. To balance efficiency and contextual richness, we retained text passages with token lengths from 500 to 2000.

For genre normalization~\citep{schreiber2015improving}, all genre labels were converted to lowercase, and spaces, hyphens (-), and slashes (/) were removed. We used the 48 root genres defined in \citep{schreiber2016genre} as our initial genre taxonomy. After filtering to retain data corresponding to the 300 most frequent genres, each was mapped to one of 20 final genre labels.

To extract artists’ regional information, we provided the first paragraph (abstract) of each Wikipedia page to Llama 3.1 8B Instruct~\citep{grattafiori2024llama}, prompting it to identify the country associated with each artist. The list of valid countries was obtained from the \textit{pycountry}\footnote{\url{https://pypi.org/project/pycountry/}} library.




\subsubsection{Data Selection}
We selected a diverse set of 500 artists based on their \textit{topic}, \textit{genre}, \textit{debut year}, and \textit{country}. Among these factors, \textit{country} was given the highest priority, with a focus on including artists from underrepresented regions. Specifically, we applied an inverse frequency strategy on MusWikiDB, prioritizing countries with lower representation to ensure geographical balance. This approach enabled the inclusion of artists from minor regions such as Mauritius, Senegal, and Azerbaijan.

\begin{figure}[!t]
\begin{center}
\includegraphics[width=\columnwidth]{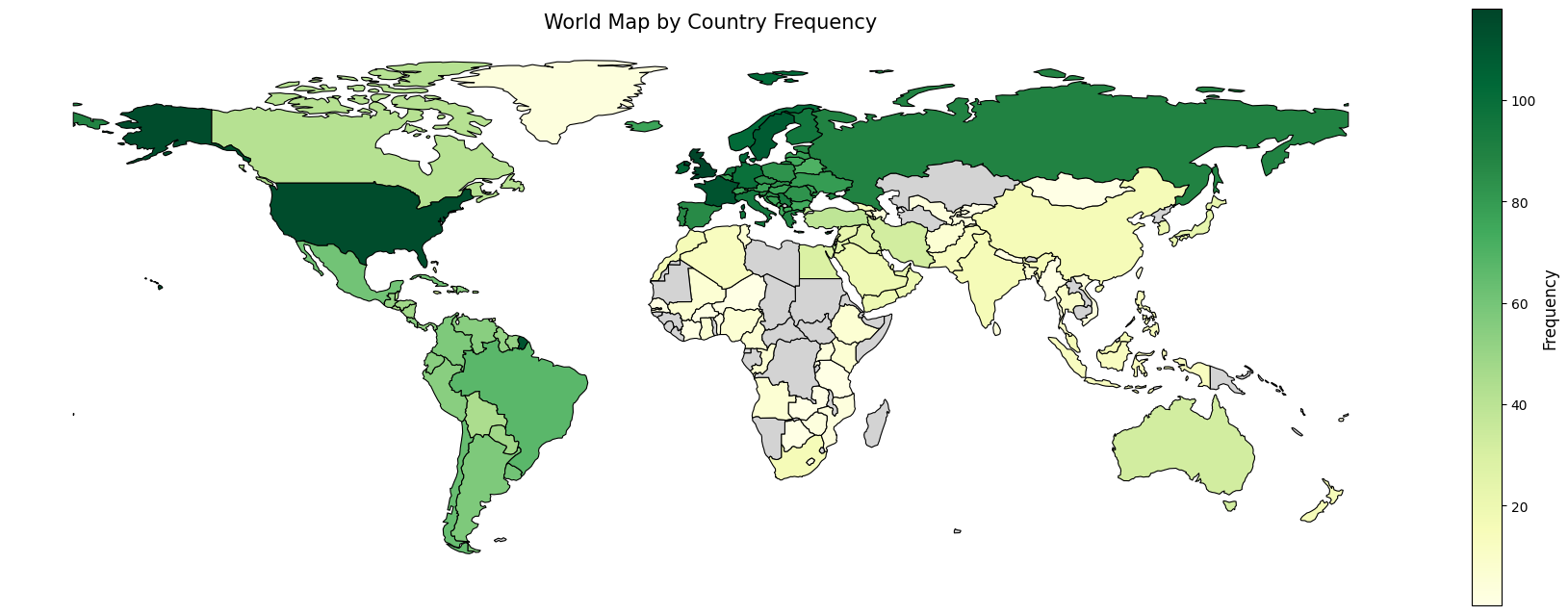}
\caption{Regional distribution of the 500 music artists in ArtistMus, spanning 163 countries (or regions) to ensure global diversity beyond the traditional U.S.- and Europe-centric focus.}
\label{fig:country_diversity}
\end{center}
\end{figure}

Figure~\ref{fig:country_diversity} illustrates the regional distribution of the artists included in ArtistMus.\footnote{You can see larger figure at Figure~\ref{fig:country_diversity_large}.} Unlike many existing datasets that are heavily biased toward the U.S. and Europe, ArtistMus deliberately incorporates artists from 163 countries (or regions), ensuring broad global diversity and representation. Approximately 43\% of the artists originate from outside the U.S. and Europe, indicating balanced representation beyond Western and European cultural centers.

\begin{figure}[!t]
\begin{center}
\includegraphics[width=\columnwidth]{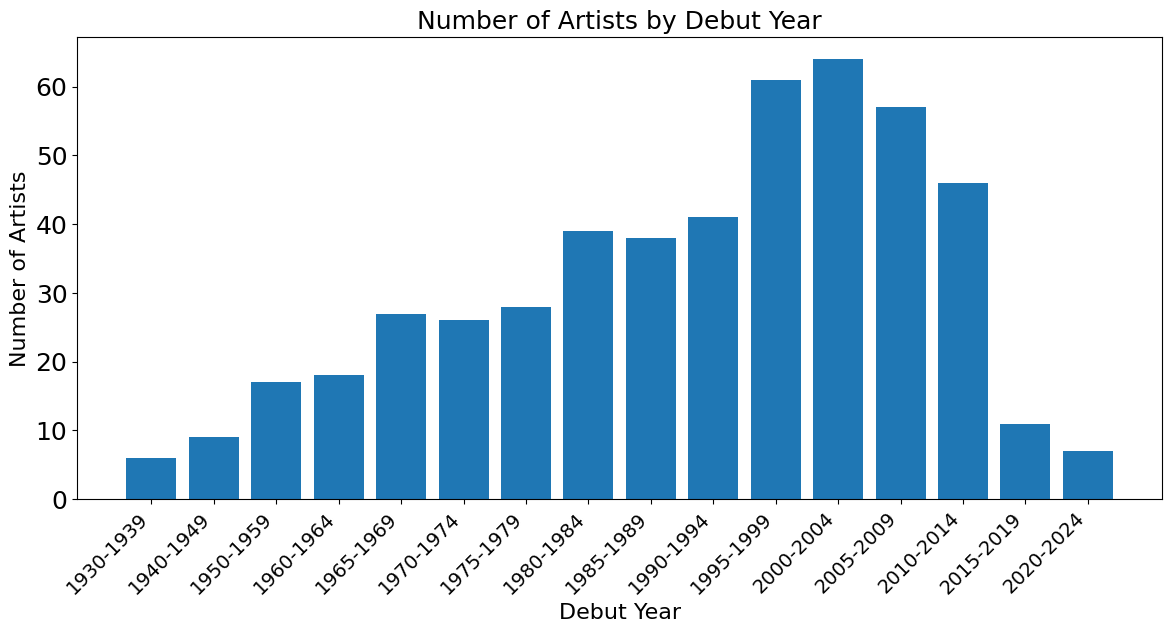}
\caption{Debut years of the 500 music artists in ArtistMus.}
\label{fig:debut_year}
\end{center}
\end{figure}

To further enhance diversity and balance, we replaced dominant genres and topics with less common ones. Instead of focusing on widely represented genres such as \textit{Pop} or \textit{Rock}, we prioritized underrepresented genres such as \textit{World Music} and \textit{Latin}. Similarly, rather than overrepresented topics like \textit{Biography}, we emphasized less frequent ones such as \textit{Artistry}. In addition, debut years were selected to achieve a more uniform temporal distribution. The distribution of debut years for the artists in ArtistMus is shown in Figure~\ref{fig:debut_year}.

For each artist, we generated one factual and one contextual question to evaluate the LLM's factual accuracy and contextual reasoning ability. To construct these questions, we provided GPT-4o~\citep{achiam2023gpt} with the corresponding section text. Factual questions target verifiable details such as dates, names, and events, whereas contextual questions require reasoning or synthesis across multiple pieces of information within the passage.

\subsubsection{Data Validation}
To ensure benchmark quality, we applied a two-stage validation process. We validate the generated questions based on two criteria: \textit{Music Relevance} and \textit{Faithfulness}. For Music Relevance, questions that did not pertain to musical aspects were excluded except important details such as the artist's birthplace. For Faithfulness, we prompted the LLM to verify whether the question and answer could be derived from the provided text. 
Finally, 1,000 multiple-choice questions passing final human validation were generated. We adjusted the placement of correct answers to ensure an even distribution, assigning 250 instances to each option.

In addition to the question–answer pairs, ArtistMus provides richer annotations, including the artist identity, topic category, the supporting text passage, as well as metadata such as genre and debut year. This richer structure enables fine-grained evaluation, for example analyzing performance across genres, eras, or topical categories.

\section{Experiments}

\subsection{Benchmarks}

\noindent For evaluation, we used two datasets: ArtistMus (in-domain) and TrustMus (out-of-domain)~\citep{ramoneda2024role}.~\footnote{Performance on the MusicTheoryBench~\citep{yuan2024chatmusician} and ZIQI-Eval~\citep{li2024music} benchmarks will be included in the camera-ready version.} Performance on factual and contextual questions was separately measured on the ArtistMus. For TrustMus, evaluation was conducted across four categories: People, Instrument \& Technology, Genre, Forms, and Theory, and Culture \& History, each comprising 100 questions. All evaluations use a multiple-choice QA format, and accuracy (exact match of the predicted option A, B, C, or D) was adopted as the evaluation metric.






\subsection{Models and Experimental Setup}
We evaluate MQA performance across two complementary experimental settings.
All experiments were conducted in a deterministic setting with temperature set to 0.

\vspace{1mm} \noindent \textbf{Main Experiments.}~~
Main experiments measure zero-shot, RAG, and Rerank performance across a diverse set of LLMs. Specifically, we evaluate:
\begin{itemize}
    \item \textbf{Zero-shot}: GPT-4o~\citep{achiam2023gpt} (API-based), Gemini 2.5 Flash (API-based), ChatMusician~\citep{yuan2024chatmusician} (music-specific), Llama 3 (1B/3B/8B)~\citep{grattafiori2024llama}, and Qwen3 (0.6B/1.7B/4B/8B)~\citep{yang2025qwen3}. For Qwen3 models, we disabled the thinking mode to ensure consistent behavior.
    \item \textbf{RAG}: Each model is equipped with first-stage retrieval using BM25 over MusWikiDB, using 256-token passages as identified as optimal in Section~\ref{sec:retriever_configuration}.
    \item \textbf{Rerank}: We add a second-stage reranking step with the BGE reranker~\citep{bge_embedding}, selecting the most relevant passages for generation.
\end{itemize}
This setup enables a systematic comparison across both closed-source and open-source models of varying scales, as well as a music-specific model, under different retrieval strategies.

\vspace{1mm} \noindent \textbf{Ablation Study.}~~
To disentangle the impact of adaptation strategies, we further compare three training paradigms on a fixed base model (Llama 3.1 8B Instruct):
\begin{itemize}
    \item \textbf{Zero-shot baseline}: No adaptation.
    \item \textbf{QA fine-tuning}: Fine-tuned on 8K multiple-choice QA pairs from MusWikiDB.
    \item \textbf{RAG-style fine-tuning}: Fine-tuned on 8K augmented \textit{(context, question, answer)} triples, enabling the model to leverage external passages during training.
\end{itemize}
This ablation isolates the effects of standard QA fine-tuning versus RAG-style fine-tuning, complementing the experiment by highlighting the benefits of incorporating contextual grounding into training.

We adopt the following training configuration for both QA fine-tuning and RAG-style fine-tuning: All fine-tuned models were trained for one epoch using LoRA~\citep{hu2022lora} with 8-bit quantization. 
We set the following hyperparameters: batch size = 2, gradient accumulation steps = 4, learning rate = 3e-5, weight decay = 0.005, warmup ratio = 0.1, cosine learning rate scheduler~\citep{wolf-etal-2020-transformers}, AdamW optimizer~\citep{loshchilov2017decoupled}, LoRA rank $r=16$, scaling factor $\alpha=16$, and dropout = 0.1. 

\subsection{Retriever Configurations}\label{sec:retriever_configuration}
We conducted an additional study on the ArtistMus benchmark to determine the optimal retriever configuration for MusWikiDB as followings:

\begin{itemize}
    \item \textbf{Passage size:} 128, 256, or 512 tokens.
    \item \textbf{Embedding models:} BM25~\citep{robertson1994some} (sparse), Contriever~\citep{izacard2021unsupervised} (dense).
    \item \textbf{Reranker:} bge-reranker-large~\citep{bge_embedding}.
\end{itemize}
For a fair comparison, we fixed the total token budget to 1024 by adjusting the number of retrieved passages: top-8 for 128-token passages, top-4 for 256, and top-2 for 512.

\section{Results}

\subsection{Main Experiments: Zero-shot vs. RAG vs. Reranker} 
\begin{table*}
\centering
\renewcommand{\arraystretch}{1.1}
\resizebox{0.95\linewidth}{!}{
\begin{tabular}{@{}ccc|cc|cc|cc@{}}
\toprule
\multicolumn{2}{l}{}                                                                                                  & \multicolumn{1}{l|}{} & \multicolumn{2}{c|}{\textbf{Zero-shot}} & \multicolumn{2}{c|}{\textbf{RAG}}      & \multicolumn{2}{c}{\textbf{Rerank}}    \\ \midrule
\multicolumn{1}{l}{}                                         & \multicolumn{1}{c|}{\textbf{Model}}                    & \textbf{Params}    & \textbf{Fact.}  & \textbf{Context.} & \textbf{Fact.} & \textbf{Context.} & \textbf{Fact.} & \textbf{Context.} \\ \midrule
\multicolumn{1}{c|}{\multirow{2}{*}{\textbf{Closed-source}}} & \multicolumn{1}{c|}{\textbf{GPT-4o}}                   & N/A                   & 67.4              & 93.0                & 93.0             & 93.0                & 95.4             & 94.0                \\
\multicolumn{1}{c|}{}                                        & \multicolumn{1}{c|}{\textbf{Gemini 2.5 Flash}}         & N/A                   & 60.4              & 93.2                & 86.6             & 90.0                & 94.2             & 91.0                \\ \midrule
\multicolumn{1}{c|}{\textbf{Music-specific}}                 & \multicolumn{1}{c|}{\textbf{ChatMusician}}             & 7B                    & 26.8              & 72.2                & 32.4             & 39.4                & 36.4             & 47.4                \\ \midrule
\multicolumn{1}{c|}{\multirow{7}{*}{\textbf{Open-source}}}   & \multicolumn{1}{c|}{\multirow{3}{*}{\textbf{Llama 3}}} & 1B                    & 29.6              & 75.0                & 62.4             & 85.4                & 66.6             & 86.2                \\
\multicolumn{1}{c|}{}                                        & \multicolumn{1}{c|}{}                                  & 3B                    & 35.4              & 83.4                & 82.8             & 91.4                & 90.0             & 93.0                \\
\multicolumn{1}{c|}{}                                        & \multicolumn{1}{c|}{}                                  & 8B                    & 39.0              & 84.8                & 85.4             & 85.6                & 92.2             & 91.2                \\ \cmidrule(l){2-9} 
\multicolumn{1}{c|}{}                                        & \multicolumn{1}{c|}{\multirow{4}{*}{\textbf{Qwen 3}}}  & 0.6B                  & 30.6              & 75.2                & 70.0             & 84.8                & 75.8             & 87.2                \\
\multicolumn{1}{c|}{}                                        & \multicolumn{1}{c|}{}                                  & 1.7B                  & 32.0              & 78.0                & 79.2             & 87.4                & 86.8             & 89.4                \\
\multicolumn{1}{c|}{}                                        & \multicolumn{1}{c|}{}                                  & 4B                    & 34.8              & 84.6                & 83.2             & 88.8                & 90.4             & 92.4                \\
\multicolumn{1}{c|}{}                                        & \multicolumn{1}{c|}{}                                  & 8B                    & 35.0              & 85.8                & 84.8             & 90.0                & 91.8             & 91.8                \\ \bottomrule
\end{tabular}
}
\caption{Performance on the ArtistMus benchmark across models and retrieval strategies. RAG uses BM25-based retrieval (top-4 passages, 256 tokens). Rerank applies BGE reranker to refine passage selection. Factual questions (Fact.) test entity recall (e.g., names, dates); contextual questions (Context.) require reasoning over narrative information. All results are accuracy (\%) on 500 questions per type.}
\label{tab:main_result}
\vspace{-0.6em}
\end{table*}

\begin{table}[t]
\centering
\resizebox{\linewidth}{!}{
\begin{tabular}{@{}c|c|c@{}}
\toprule
\textbf{Models}                   & \textbf{\# Params} & \textbf{Zero-shot / RAG / Rerank} \\ \midrule
\multirow{3}{*}{\textbf{Llama 3}} & 1B                 & 0.25               / 0.36         / 0.38  \textcolor{OliveGreen}{\textbf{\footnotesize (+13\%)}}          \\
                                  & 3B                 & 0.29               / 0.38         / 0.39  \textcolor{OliveGreen}{\textbf{\footnotesize (+10\%)}}          \\
                                  & 8B                 & 0.36               / 0.41         / 0.43   \textcolor{OliveGreen}{\textbf{\footnotesize (+7\%)}}         \\ \midrule
\multirow{4}{*}{\textbf{Qwen 3}}  & 0.6B               & 0.28               / 0.36         / 0.38      \textcolor{OliveGreen}{\textbf{\footnotesize (+10\%)}}      \\
                                  & 1.7B               & 0.33               / 0.39         / 0.39     \textcolor{OliveGreen}{\textbf{\footnotesize (+6\%)}}       \\
                                  & 4B                 & 0.32               / 0.38         / 0.41    \textcolor{OliveGreen}{\textbf{\footnotesize (+9\%)}}        \\
                                  & 8B                 & 0.34               / 0.40         / 0.43    \textcolor{OliveGreen}{\textbf{\footnotesize (+9\%)}}        \\ \bottomrule
\end{tabular}
}
\caption{Performance on the TrustMus benchmark. In this setting, RAG corresponds to first-stage retrieval based on BM25, whereas Rerank represents two-stage retrieval employing the BGE reranker.}
\label{tab:trustmus_result}
\vspace{-1em}
\end{table}

\paragraph{In-domain (ArtistMus).}  
Table~\ref{tab:main_result} summarizes the results of a wide range of LLMs under three inference settings: zero-shot, RAG, and RAG+Reranker.
Across all models, factual questions were notably more challenging than contextual ones. For example, GPT-4o outperformed Llama 3.1 8B by 28.4 pp on factual questions in zero-shot setting, while the gap narrowed to 8.2 pp on contextual questions. RAG substantially improved factual accuracy across open-source models: Llama 3.1 3B improved from 39.0\% to 85.4\% (+46.4 pp), and Qwen3 8B from 35.0\% to 84.8\% (+49.8 pp). Adding a reranker further boosted accuracy, with gains of up to +7.6 pp compared to RAG. These results confirm that retrieval is more beneficial for entity-centric questions such as artist names and works.

GPT-4o and Gemini 2.5 Flash achieved the strongest overall performance, but even smaller open-source models (e.g., Llama 3.1 3B, Qwen3 1.7B) benefited substantially from RAG and reranking. With retrieval support, these smaller models approached the performance of much larger API-based models, demonstrating that RAG can effectively bridge the gap between open-source and proprietary systems. ChatMusician, despite being designed for music-specific tasks, achieved only around 26\% accuracy—comparable to random guessing—largely due to frequent instruction-following failures. Similarly, the RAG setting exhibited degraded contextual accuracy, primarily because the models often failed to properly follow task-specific instructions when integrating retrieved information.

\paragraph{Out-of-domain (TrustMus).}
Table~\ref{tab:trustmus_result} presents results on the TrustMus benchmark, which represents a different knowledge source than Wikipedia. Despite this shift, our methodology continued to provide consistent gains: for instance, Llama 3.1 1B improved from 25.0\% to 36.0\% (+11.0 pp) with RAG, and further to 38.0\% (+2.0 pp) with reranking. This indicates that retrieval with MusWikiDB not only strengthens in-domain factual recall but also transfers effectively to out-of-domain knowledge, even when the benchmark corpus differs from the retrieval database.

\paragraph{Analysis.} The substantial factual accuracy gains from RAG (+40.4 pp on average) contrast with modest contextual improvements (+4.8 pp on average), revealing an important distinction~\footnote{ChatMusician is excluded because of low instruction following rate.}. Factual questions often require precise entity recall (e.g., album names) that are absent from model parameters but directly retrievable from passages. Contextual questions, which demand reasoning over information, benefit less from retrieval alone—models already possess adequate reasoning capabilities once factual grounding is provided. This suggests a division of labor: retrieval provides facts, while model parameters handle reasoning. The smaller gap between model sizes with RAG (e.g., Qwen 0.6B vs 8B: 77.4\% vs 87.4\% on average) further indicates that retrieval mitigates the need for large parametric memory, democratizing access to factual knowledge.

\begin{table}[t]
\centering
\resizebox{\linewidth}{!}{
\begin{tabular}{@{}l|cc|c@{}}
\toprule
\multicolumn{1}{l|}{}  & \textbf{Factual} & \textbf{Contextual} & \textbf{Average} \\ \midrule
\textbf{Zero-shot}     & 39.0             & 84.8                & 61.9             \\
\textbf{QA FT}         & 40.8             & 75.6                & 58.2             \\
\textbf{RAG}           & 85.4             & 85.6                & 85.5             \\
\textbf{Rerank}        & \textbf{92.2}             & 91.2                & \underline{91.7}             \\
\textbf{RAG FT}        & \underline{86.8}             & \underline{93.0}                & 89.9             \\
\textbf{RAG FT Rerank} & \textbf{92.2}             & \textbf{94.0}                & \textbf{93.1}             \\ \bottomrule
\end{tabular}
}
\caption{Ablation study. QA FT refers to the zero-shot performance of the QA fine-tuned model, RAG and Rerank are taken from the main experiment, and RAG FT and RAG FT Rerank correspond to the performance of the RAG-style fine-tuned model with RAG and reranker applied. The best result is highlighted in \textbf{bold}, while the second best is \underline{underlined}.}
\label{tab:ablation}
\vspace{-1em}
\end{table}

\subsection{Ablation Study: Fine-tuning Strategies}
Table~\ref{tab:ablation} reports the results of different fine-tuning strategies on ArtistMus using the Llama 3.1 8B Instruct model. Several observations emerge:

\begin{itemize}
    \item \textbf{QA Fine-tuning:} Compared to the zero-shot baseline (39.0\% / 84.8\%), QA fine-tuning slightly improved factual accuracy to 40.8\% (+1.8 pp) but led to a large drop in contextual performance to 75.6\% (–9.2 pp). This suggests that conventional QA fine-tuning may overfit to training data, improving memorization while harming general inference ability.
    \item \textbf{RAG Inference vs. Reranking:} RAG inference yielded a dramatic gain in factual accuracy (+46.4 pp), while maintaining contextual accuracy. Adding a reranker further boosted both factual (+6.8 pp) and contextual (+5.6 pp) performance, showing the importance of two-stage retrieval.
    \item \textbf{RAG-style Fine-tuning:} Incorporating retrieved passages during training improved contextual understanding, with accuracy rising to 93.0\% (+7.4 pp), while factual accuracy remained strong (86.8\%). With reranker, it achieved the best overall performance (93.1\%), confirming that training the model to leverage retrieved evidence enhances both factual recall and contextual reasoning.
\end{itemize}

\subsection{Retriever Configuration Study}
Table~\ref{tab:rag_chunk_embedding} shows the effect of varying passage size, embedding models, and the usage of reranker. 
\begin{itemize}
    \item Contextual accuracy generally improved with shorter passages, while factual accuracy was more stable across passage sizes for BM25 but degraded significantly for Contriever.
    \item BM25 with a reranker consistently outperformed dense retrieval methods, reflecting the entity-centric nature of ArtistMus.
    \item The best performance was achieved with BM25 + reranker at 256 tokens, balancing accuracy and efficiency.
\end{itemize}

\begin{figure}[!t]
\begin{center}
\includegraphics[width=\columnwidth]{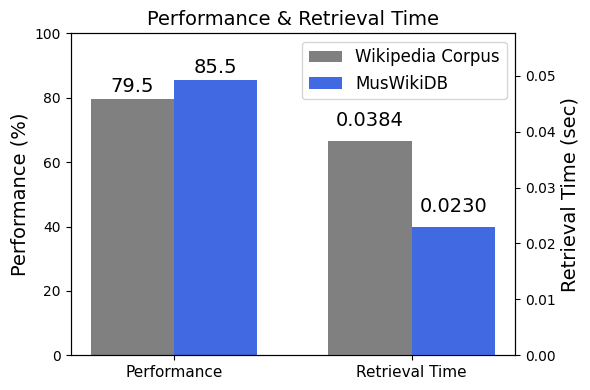}
\caption{RAG performance and retrieval time for Wikipedia Corpus~\citep{karpukhin2020dense} and MusWikiDB.}
\label{fig:db_difference}
\end{center}
\vspace{-1em}
\end{figure}

As shown in Figure~\ref{fig:db_difference}, MusWikiDB achieves both higher retrieval accuracy (+6.0 pp absolute improvement) and significantly faster (40\%) retrieval speed  compared to the general Wikipedia corpus. These results highlight the effectiveness of a music-specialized knowledge base: not only does it provide more relevant evidence for music question answering, but it also enables efficient large-scale inference, which is crucial for real-time applications.


\begin{table}[!t]
\centering
\resizebox{\linewidth}{!}{
\begin{tabular}{@{}lccc@{}}
\toprule
\textbf{Embedding}                            & \textbf{Passage Size} & \textbf{Factual} & \textbf{Contextual} \\ \midrule
\multirow{3}{*}{\textbf{  BM25}}                & 128                   & 81.2             & 87.2                \\
                                              & 256                   & 84.4             & 85.6                \\
                                              & 512                   & 85.0             & 85.8                \\ \midrule
\multirow{3}{*}{\textbf{  BM25$^R$}}       & 128                   & 85.6             & 89.6                \\
                                              & 256                   & \textbf{92.2}    & \underline{91.2}          \\
                                              & 512                   &  \underline{91.0}       & 89.8                \\ \midrule
\multirow{3}{*}{\textbf{  Contriever}}          & 128                   & 80.8             & 80.0                \\
                                              & 256                   & 69.2             & 84.6                \\
                                              & 512                   & 54.6             & 75.4                \\ \midrule
\multirow{3}{*}{\textbf{  Contriever$^R$}} & 128                   & 86.8             & 82.0                \\
                                              & 256                   & 84.2             & \textbf{91.6}       \\
                                              & 512                   & 70.6             & 87.6                \\ \bottomrule
\end{tabular}
}
\caption{Llama 3.1 8B Instruct~\citep{grattafiori2024llama} RAG performance on ArtistMus, by different passage size and embeddings. $^R$ indicates the use of a reranker. The best result is highlighted in \textbf{bold}, while the second best is \underline{underlined}.}
\label{tab:rag_chunk_embedding}
\end{table}


\section{Conclusion}

This paper introduces two key contributions for music information retrieval: MusWikiDB, a domain-specific knowledge base with over 3M passages from 144K music-related Wikipedia pages, and ArtistMus, a benchmark of 1000 questions covering 500 globally diverse artists. Together, they fill critical gaps in music question answering (MQA).

Our experiments show that retrieval augmented generation (RAG) substantially boosts factual accuracy and overall robustness in MQA, narrowing the gap between open-source and proprietary systems. Notably, smaller models equipped with retrieval can match or even surpass large closed models, demonstrating that domain-specific retrieval democratizes access to high-quality, knowledge-grounded MQA. MusWikiDB also enables more accurate and efficient retrieval than the general Wikipedia corpus, validating the value of domain-specialized corpora for real-time applications.

Furthermore, integrating retrieval into model training enhances both factual and contextual reasoning, establishing a new paradigm for unifying retrieval and generation in music-domain LLMs. The consistent gains observed across in-domain and out-of-domain benchmarks highlight the transferability of this approach. 


\section{Ethics Statement}

MusWikiDB is constructed from Wikipedia, which, while comprehensive, reflects the biases inherent in crowdsourced knowledge bases, including underrepresentation of non-Western music traditions and artists from marginalized communities. While ArtistMus includes artists from 163 countries, the quality and depth of available information varies significantly by region and genre. 

Although RAG substantially improves factual grounding, our system may still generate incorrect information, particularly for less-documented artists or culturally contested interpretations. We emphasize that generated outputs should be treated as starting points for exploration rather than authoritative sources, especially in applications where misinformation could impact artist reputations or cultural understanding.

This work is designed to advance research in music information retrieval and domain-specific QA systems. We caution against deploying such systems in high-stakes applications without human oversight and appropriate disclaimers. The benchmark and resources are released to enable further research into addressing these challenges.

\section{Limitations}


ArtistMus focuses on artist-centric questions with a multiple-choice format. While this enables consistent automatic evaluation, it does not capture all aspects of music question answering, such as open-ended explanations, comparative analysis across artists, or questions requiring multi-hop reasoning. The 1,000-question scale, though sufficient for establishing trends, could be expanded to enable more fine-grained analysis across subgenres and eras.

MusWikiDB relies exclusively on Wikipedia content, which provides broad coverage but may lack depth for niche genres, regional music traditions, or recently emerging artists. The crawling strategy based on hyperlink depth may also introduce coverage gaps. Future work could integrate additional structured music databases to provide more comprehensive and up-to-date information.

\newpage
\section{Bibliographical References}\label{sec:reference}

\bibliographystyle{lrec2026-natbib}
\bibliography{lrec2026}

\newpage
\appendix
\section{Appendix}

\subsection{Examples of TrustMus}

\begin{tcolorbox}[colback=gray!5, colframe=black!50,
                  title=\textbf{Category: People}, left=5pt]

\textbf{Question:} \\
"When did Adolph Martin Schlesinger found the music-publishing house?" \\
A. In 1819 \\
B. Before 1795 \\
C. In August 1814 \\
D. In April 1810 \\
\textbf{Answer:} D
\end{tcolorbox}

\begin{tcolorbox}[colback=gray!5, colframe=black!50,
                  title=\textbf{Category: Instrument \& Technology}, left=5pt]

\textbf{Question:} \\
"What is the name of the highland bagpipe that was brought by Scottish regiments during the British Raj?" \\
A. Tablkhāna \\
B. Bagpipe \\
C. Naqqārakhāna \\
D. Nobat \\
\textbf{Answer:} B
\end{tcolorbox}

\begin{tcolorbox}[colback=gray!5, colframe=black!50,
                  title=\textbf{Category: Culture \& History}, left=5pt]

\textbf{Question:} \\
"When did Amalarius of Metz die?" \\
A. c830 \\
B. c825 \\
C. c850 \\
D. c775 \\
\textbf{Answer:} C
\end{tcolorbox}

\begin{tcolorbox}[colback=gray!5, colframe=black!50,
                  title=\textbf{Category: Genre \& Forms \& Theory}, left=5pt]

\textbf{Question:} \\
"What did Christoph Bernhard describe as 'harmonic counterpoint' in the 17th century?" \\
A. The regulated alternation of perfect and imperfect consonances \\
B. The relationships between such sounds \\
C. The simultaneous sounding of two or three notes in isolation \\
D. An articulated sequence of 'well-juxtaposed consonances and dissonances' \\
\textbf{Answer:} D
\end{tcolorbox}

\begin{figure*}[t]
\begin{center}
\includegraphics[width=2.0\columnwidth]{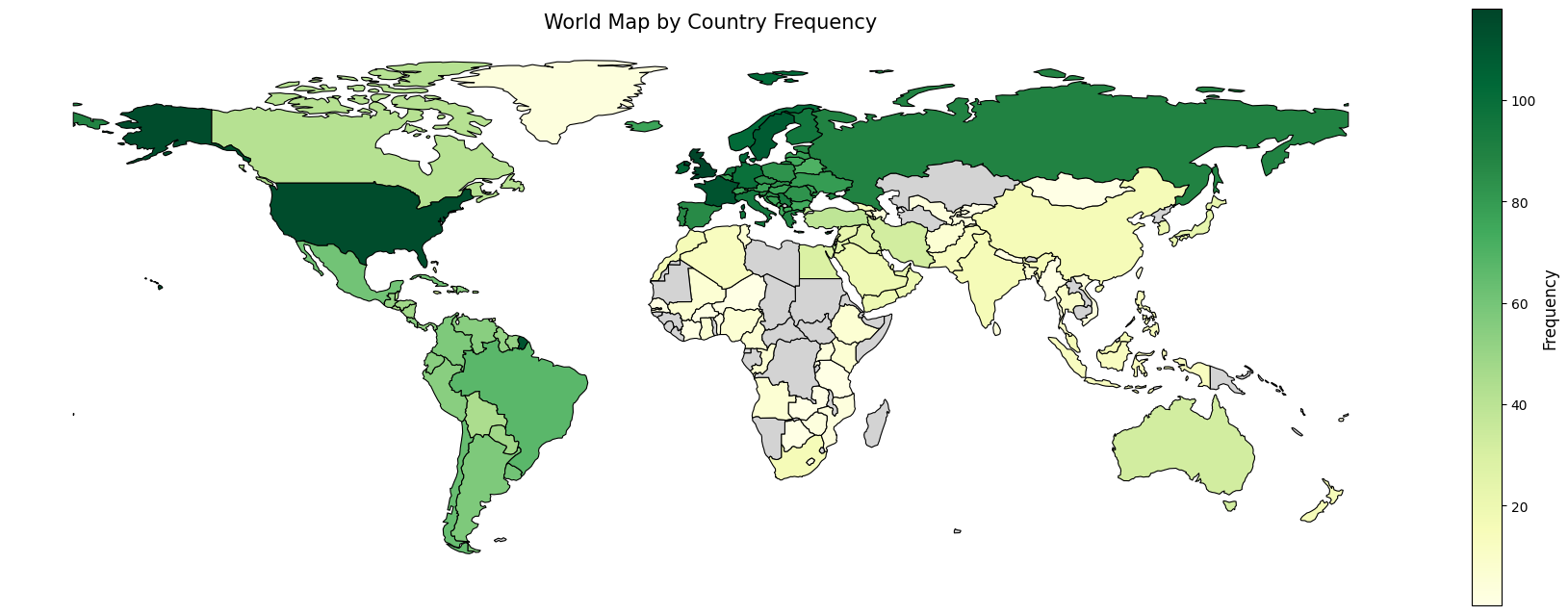}
\caption{Regional distribution of the 500 music artists in ArtistMus, spanning 163 countries (or regions) to ensure global diversity beyond the traditional U.S.- and Europe-centric focus.}
\label{fig:country_diversity_large}
\end{center}
\end{figure*}

\bibliographystylelanguageresource{lrec2026-natbib}
\bibliographylanguageresource{languageresource}

\end{document}